\documentclass[graybox]{svmult}

\usepackage{type1cm}        
\usepackage{makeidx}         
\usepackage{graphicx}        
\usepackage{multicol}        
\usepackage[bottom]{footmisc}
\usepackage{color}
\usepackage{colortbl}
\usepackage[colorlinks]{}
\usepackage{wrapfig,lipsum,booktabs}
\usepackage[table]{xcolor}
\usepackage{newtxtext}       %
\usepackage{newtxmath}       

\definecolor{burntorange}{rgb}{0.8, 0.33, 0.0}
\definecolor{light-gray}{gray}{0.9}
\definecolor{richelectricblue}{rgb}{0.03, 0.57, 0.82}
\definecolor{brightyellow}{rgb}{255,255,0}
\definecolor{lemonchiffon}{rgb}{255,250,205}
\definecolor{papayawhip}{rgb}{255,239,213}
\definecolor{moccasin}{rgb}{255,228,181}

\makeindex             

\begin{document}

\title*{Jurassic is (almost) All You Need: \\Few-Shot Meaning-to-Text Generation for Open-Domain Dialogue}
\titlerunning{ Few-Shot Meaning-to-Text Generation for Open-Domain Dialogue} 
\author{Lena Reed, Cecilia Li, Angela Ramirez, Liren Wu, and Marilyn Walker}
\authorrunning{Reed, Li, Ramirez, Wu and Walker}
\institute{University of California Santa Cruz, Ca.  95062
\email{mawalker@ucsc.edu}}
%
%
\maketitle

\vspace{-.3in}
\abstract{One  challenge with  open-domain dialogue systems is the need to 
produce truthful, high-quality responses on any topic.   We aim to improve the quality and  coverage of Athena, an  Alexa Prize dialogue system. We experiment with few-shot prompt-based learning, comparing  GPT-Neo to Jurassic-1, for the movies, music, TV, sports, and  video game domains, both within and cross-domain, with different prompt set sizes (2, 3, 10),  formats, and  meaning representations consisting of either sets
of WikiData KG triples, or dialogue acts. 
Our evaluation uses {\sc bleurt}  and human  metrics,
and shows that with 10-shot prompting,  Athena-Jurassic's performance is significantly better 
for coherence and semantic accuracy. Experiments with 2-shot cross-domain prompts results in
a huge performance drop for Athena-GPT-Neo, whose semantic accuracy falls to 0.41, and whose untrue
hallucination rate increases to 12\%. Experiments with dialogue acts for video games show that with 10-shot prompting, both models learn to control dialogue acts, but Athena-Jurassic has significantly higher coherence, and only 4\%  untrue
hallucinations. Our results suggest that Athena-Jurassic  produces
 high enough quality outputs to be useful in live systems with real users. To our knowledge, these are the first results demonstrating  that few-shot semantic prompt-based learning can create NLGs that generalize to new domains,
and produce high-quality, semantically-controlled, conversational responses directly from meaning representations.}

\section{Introduction}
\label{sec:intro}

One challenge with open-domain dialogue systems is the need to respond to users' utterances on any topic with high-quality responses. To handle this challenge, a common approach is to use an ensemble of response generators (RGs) and then train a ranker to select from a pool of possible responses \cite{curry2018alana,shalyminov2018neural,harrison2020athena,paranjape2020neural,chen2018gunrock,fang2018sounding18,gabriel2020further}. The ensemble of RGs can use a variety of generation techniques. One type of RG generates responses directly from the dialogue context, using a pre-trained model such as GPT-2 that is possibly tuned on additional conversational data, such as Topical Chat or Empathetic Dialogues\cite{gopalakrishnan2019topical,paranjape2020neural}. Knowledge-Grounded response generation is a variant of this approach, where knowledge in the form of text is  available during generation to control the utterance content and veracity\cite{hedayatnia2020policy,yavuz2019deepcopy}. Template-Based RGs are also quite common, where templates are hand-written and then populated with values from a database or knowledge graph (KG). 

We are primarily interested in Meaning-to-Text (M2T) NLG engines, where the desired meaning is specified and the NLG engine produces one or more utterances that express that meaning. These can be trained with parallel corpora of meaning representations (MRs) and reference utterances \cite{colin2016webnlg,Novikovaetal17,walker2018exploring,budzianowski2018multiwoz,harrison2019maximizing,tran2017neural}, or can be trained by tuning  a large language model  with a domain specific training set ~\cite{juraskawalker21,zhou2020design,zhao2020bridging,oraby2019curate,reedetal20}. These M2T engines are most often used for task-oriented responses on specific (sets of) topics, such as restaurants, sports, movies,  hotels, laptops, TVs, and video games ~\cite{gardent2017creating,KedzieMcKeown19,eric2019multiwoz,cervone2019natural}. 

\begin{wrapfigure}{r}{2.7in}
    \vspace{-.2in}
    \small
    \def\arraystretch{1.15}
    \begin{tabular}{p{0.2in} >{\raggedright\arraybackslash}p{0.92\linewidth}}
     \hline
      A1: &  Do you have a favorite musician? \\
      \rowcolor{light-gray}
      U1: &  yeah, taylor swift. \\ \hline
      A2: & Ok, Taylor Swift. \textcolor{burntorange}{\textbf{Wow!  Taylor Swift is very prolific!
        She has  114 songs, that's a lot!}} \\
       & (Taylor Swift, numTracks, 114)  \\ \hline
       \rowcolor{light-gray}
        U2:  &  yeah, that is a lot \\ \hline
     A3: &   \textcolor{burntorange}{\textbf{I like Taylor Swift's song, 22. It came out in 2013.}} 
        Do you like that song? \\
       &(Taylor Swift, song, 22), (22, date, 2013) \\ \hline
        \rowcolor{light-gray}
        U3: &  yeah, it's a good one. \\ \hline
     A4:  &    Right? This is interesting, \textcolor{burntorange}{\textbf{Taylor Swift sings the 
        song Bad Blood with Kendrick Lamar}}, want to hear more about
        Kendrick Lamar? \\
        & (Bad Blood, performer, Taylor Swift),  (Bad Blood, performer, Kendrick Lamar)
        \\ \hline
        \rowcolor{light-gray}
        U4: &  not really \\ \hline
        A5:  & One of my favorite musicians is P!nk. Isn't this
        interesting, \textcolor{burntorange}{\textbf{P!nk won a Grammy Award for Best Pop Collaboration
        with Vocals}}. Do you want to hear more? \\
        & (P!nk, award, Grammy Award for Best Pop Collaboration
        with Vocals) \\ \hline
    \end{tabular}
    \caption{A conversation about music between Athena (A) and a user (U).
    The music KG-RG interweaves template-based utterances based on  KG triples with
    other template-based utterances.  Response segments generated using the KG triples  are highlighted in orange. For each segment, the KG triples used  by the RG as the input MR are  shown below the response.}
    \label{fig:music_kg}
    \vspace{-0.2in}
\end{wrapfigure}
Recent work on language models as few-shot learners \cite{brown2020language} suggests a novel paradigm for creating a M2T NLG engine with prompt-based learning, where prompts are provided that condition the language model  to produce a certain type of output. In our case, these conditioning prompts consist of  pairs of meaning representations and their corresponding output. Our  aim in this paper is to investigate how prompt-based learning can improve the quality and the coverage of the M2T RGs in Athena, a dialogue system that has been a finalist in the Alexa Prize for the last two years~\cite{harrison2020athena,patilathena2021}. 

There are two types of M2T RGs in Athena. One set are template-based whose MRs are  sets of triples  from the WikiData Knowledge-Graph (KG-RGs). These KG-RGs cover the movies, sports, music, and TV topics. One advantage of using WikiData is the automatic updates to its knowledge by its editors. Another benefit is the ability to traverse the KG to new relations or entities, to say more things about a topic in a dialogue \cite{moon2019opendialkg}. A third benefit is that Athena represents named entities and  pronouns with  their Wikidata IDs, providing a direct link to the KG \cite{patilathena2021}. Figure~\ref{fig:music_kg} provides
a conversation with the Music KG-RG  that shows how the  knowledge triples from WikiData are realized with templates.\footnote{In accordance with the Alexa Prize  rules, the shared conversations in  Figure~\ref{fig:music_kg} and Figure~\ref{viggo-conv} are between Athena and our team, or UCSC undergraduates, rather than real users.}  
The other RG, for the video games topic, is based on the Viggo corpus  \cite{juraska2019viggo}.
This is a parallel corpus of MRs and crowd-sourced realizations. An conversation with the Viggo RG is below in Figure~\ref{viggo-conv}.   

Both of these  RGs require substantial  human effort. The KG-RGs require two steps: (1) sets of \emph{interesting} and \emph{frequently populated} KG relations must be selected by hand ~\cite{moon2019opendialkg,patilathena2021} and, (2) templates must be hand-written to realize them. This means that they currently cover a limited set of relations, ones that are populated frequently enough to make writing templates worthwhile. As previous work on dialogue generation has shown, even combinations of existing relations typically require multiple additional templates to be written \cite{rambow2001evaluating,rambow2001natural,walker2002automatically}. The existing  KG-RG entities and relations are  in Table~\ref{table:kg_topics}, as well as novel KG-RG relations and entities
that we experiment with below with 2-shot prompting.

\begin{table}[htb]
    \small
    \begin{center}
   \begin{tabular}{@{}p{0.4in} |>{\raggedright\arraybackslash}p{.9in}|>{\raggedright\arraybackslash}p{3.09in} @{}} \toprule
        {\bf Topic} & {\bf Entities} & {\bf Relations} \\ \hline \hline
        {\bf Movies} & Movies Actors Directors Awards* & cast  voiceCast  
        spouse  childrenNum  genre  award  director*  work*  date*  screenWriter*  producer* \\ \hline
        {\bf Music} & Musicians~Bands Awards* Songs* Albums* & performer (song and album) numTracks
        genre  award  memberOf  instrument  label  date*  show*  work*  \\ \hline
        {\bf Sports} & Athletes Sports Awards* & team  position  participant 
        (tournament,  leagues)  spouse  childrenNum  award  height  date*  work*  ranking*  duration*  reviewScoreBy*  disciplineCompetedIn*  numMatches*  numAwards*  draftedBy*  draftPicknum*  startTime* \\ \hline
        {\bf TV} & Shows Assessment* characterRole* & cast  role  creator  director  genre  award  characterRole*  narrativeLocation*  mainSubject*  assessment*  assessmentOutcome*   hasPart*  occupation*  derivativeWork*  startTime*  endTime*   filmingLocation*  setInPeriod*  numSeasons*  numEpisodes*       \\
        \bottomrule
    \end{tabular}
    \end{center}
    \caption{The KG topics,  entities and relations  in the Athena-KG-synthetic  corpus. A * indicates novel entities and relations that are tested in Section~\ref{results-sec}.}     \label{table:kg_topics}
    \vspace{-0.2in}
\end{table}

The neural
 Viggo RG required  constructing dialogue acts and meaning representations, populating their content with information from IGDB, crowd-sourcing responses, and cleaning the resulting corpus,
in order to experiment with methods for improving the  fluency and semantic accuracy responses   \cite{juraska2019viggo,juraskawalker21}. Thus the ability to reliably generate high-quality responses directly from MRs via neural NLGs would transform the use of M2T NLGs in dialogue systems \cite{colin2016webnlg,Lebretetal16,WisemanSR17,eric2019multiwoz,rastogi2020towards,zhao2020bridging}.

\begin{figure}[t!b]
    \small
    \begin{tabular}{@{}p{0.4in} |p{3.99in} @{}} \toprule
    {\bf Topic} &   {\bf Example} \\
    \hline \hline 
 {\bf Movies}  &  {\cellcolor[gray]{0.9}(Wonder Woman, director, Patty Jenkins)}\\ 
 & I believe I read that Wonder Woman is directed by Patty Jenkins. Sometimes a director's unique voice really comes through in their work. Do you think that happened in this case? \\
    	\hline
 {\bf Music} & 	{\cellcolor[gray]{0.9}(Rihanna, record label, Def Jam Records)}	\\ 
 & Here's another musician who worked for the same label Def Jam Records, called Rihanna. Want to hear about them? \\ \hline
{\bf Sports}	& {\cellcolor[gray]{0.9} (Lebron James, member of sports team, Los Angeles Lakers), (Lebron James, position played on team/specialty, power forward)}	\\ 
 & Lebron James has played on many famous teams such as the Los Angeles Lakers, and played many positions like power forward. 	\\ \hline 
{\bf TV} & {\cellcolor[gray]{0.9}	(Lost, genre, paranormal television program), (Lost, genre, drama television)}		\\ 
&  Lost is considered both a paranormal television program and a drama television. What's your opinion of paranormal television programs or drama television shows? \\ \hline
    \end{tabular}
  	\vspace{-0.1cm}
	\caption{Sets of KG triples used as meaning representations, and the corresponding utterances. 
	Dialogue Act types are not represented in the MRs
	but note that some  KG triples are realized with  multiple dialogue acts, combining statements, opinions and  questions.}
    \label{kg-generation-examples}
  	\vspace{-0.3cm}
\end{figure}

We utilize Athena's current RGs to create prompt and test sets  for  two  new neural Meaning-to-Text RGs, Athena-GPT-Neo and Athena-Jurassic \cite{black2021gpt,lieberetal21,brown2020language}. 
We conduct 
few-shot  prompt-based learning experiments, where we systematically vary within and cross-domain prompts, different prompt set sizes (2, 3, 10),   prompt formats, and  type of meaning representations.  
We expect that these NLGs to generalize beyond
their conditioning data ~\cite{radford2019language,ham2020end,budzianowski2019hello,pang-etal-2020-towards}.
 We evaluate
the results using both {\sc bleurt}  and human evaluation. Our results show that, with 10-shot conditioning, both Athena-GPT-Neo and Athena-Jurassic generally produce coherent outputs, but that for within-domain experiments, Athena-Jurassic's performance is significantly better 
for the human evaluation metrics
of  coherence and semantic accuracy. Experiments with 2-shot prompts on completely novel MRs results in a
huge performance drop for Athena-GPT-Neo, whose semantic accuracy falls to 0.41, and  untrue
hallucination rate  increases to 12\%. Experiments with the Viggo video games corpus shows that, with 10-shot
prompts, both Athena-GPT-Neo and Athena-Jurassic can learn to control the dialogue acts realized, but Athena-Jurassic has significantly higher coherence, mainly because Athena-GPT-Neo
produces some redundant and repetitive utterances.  Athena-GPT-Neo also produces  untrue
hallucinations in 12\% of the video game outputs. We use  the human evaluation  to examine whether the {\sc bleurt} scores are meaningful with results showing that {\sc bleurt} scores have a very good correlation
with semantic accuracy, but not with coherence, or other human metrics.
Our results suggest that Athena-Jurassic can reliably produce outputs that are high enough quality to be used in  live systems with real users.
To our knowledge, this is the first paper
to show  that few-shot prompt-based learning  can create NLGs that generalize to new domains,
and
produce high-quality, semantically-controlled, conversational responses directly from MRs and KG triples.

\section{Datasets}
\label{data-sec}

We use two datasets for prompt-based learning for Athena's response generators. One is based on Athena's KG RGs for the domains of music, movies, sports, and TV. Figure~\ref{kg-generation-examples} provides KG triples (entity, relation, entity) and responses produced by the current KG-RGs. 
Each response is designed to continue the conversation with one or more statements, opinions, and/or question dialogue acts about an entity. However, the  MRs represent the triples and do not explicitly represent dialogue acts.
Some responses are based on traversing the KG links to a related entity, and introducing that entity, as illustrated in the music example about {\it Rihanna}, where the record label {\it Def Jam Records} is used to link two music artists. A music conversation was given in Figure~\ref{fig:music_kg} above. Both existing and novel KG-RG entities and relations are  in Table~\ref{table:kg_topics}. 

\begin{wrapfigure}{r}{3.0in}
 \vspace{-.2in}
    \small
   	\centering
   	\def\arraystretch{1.15}
    \begin{tabular}{p{0.2in} >{\raggedright\arraybackslash}p{2.8in}}
		\hline
    	A1:  & \dots Well, one of my all-time favorites is Control. I have already completed it twice! Have you played it by any chance? \\
    	\hline
    	\rowcolor{light-gray}
    	U1:  & no not yet \\
    	\hline
    	A2:  & Obviously. There are just too many fun games for one person to play in a single lifetime. At any rate, \textcolor{burntorange}{\textbf{Control} is an \textbf{excellent} \textbf{third-person} \textbf{action-adventure} \textbf{shooter} with \textbf{multiplayer}. It was released in \textbf{2019} by \textbf{Remedy Entertainment} for \textbf{PC}, \textbf{PlayStation}, and \textbf{Xbox}.} Do you think you would potentially like it? \\
    	& \textcolor{darkgray}{\emph{inform}(name[\textbf{Control}], release\_year[\textbf{2019}], developer[\textbf{Remedy Entertainment}], rating[\textbf{excellent}], genres[\textbf{shooter}, \textbf{action-adventure}], has\_multiplayer[\textbf{yes}], player\_perspective[\textbf{third person}],  platforms[\textbf{PC}, \textbf{PlayStation}, \textbf{Xbox}])} \\
    	\hline
    	\rowcolor{light-gray}
    	U2:  & absolutely i was actually hoping to play it sometime soon \\
    	\hline
    	A3:  & Alright! \textcolor{burntorange}{Have you played any \textbf{adventure} games by \textbf{Remedy Entertainment}, like \textbf{Alan Wake}?} \\
    	& \textcolor{darkgray}{\emph{suggest}(name[\textbf{Alan Wake}], developer[\textbf{Remedy Entertainment}], genres[\textbf{adventure}])} \\
    	\hline
    	\rowcolor{light-gray}
    	U3:  & yes i did i loved the dark atmosphere in alan wake \\
    	\hline
    	A4:  & Yeah! I remember playing Alan Wake some time ago, and it wasn't bad at all. \textcolor{burntorange}{Do you like \textbf{multiplayer} games?} \\
    	& \textcolor{darkgray}{\emph{request\_attribute}(has\_multiplayer[])} \\
    	\hline
    \end{tabular}
 	\vspace{0.1cm}
	\caption{A conversation between Athena(A) and a user (U) about video games, exhibiting a smooth interweaving of neural and template-based responses. Response segments generated by the neural model are highlighted in orange. For each such segment, the MR constructed by the RG as input for the model is shown below the response.}
    \label{viggo-conv}
 	\vspace{-0.2in}
 \end{wrapfigure}
To use prompt-based learning to create new  KG-RGs, we create a new corpus, Athena-KG-Synthetic, of sets of knowledge triples and their template-based responses. We select five  template categories and their paraphrases from the movies RG, two from  music,  three from  sports, and two from  TV.\footnote{Viggo and the Athena-KG-Synthetic corpus are available from nlds.soe.ucsc.edu.}  We query WikiData for thousands of KG triples to populate the templates and then split the resulting dataset into $\sim$32K train, 3558 development, and a test set of  100 instances for each template category.

The second dataset is the Viggo dataset. The Viggo RG combines responses generated from templates with those generated from meaning representations, as shown in Figure~\ref{viggo-conv}. The set of dialogue acts (DAs) are carefully constructed to be conversational and engage the user, rather than being purely informative ~\cite{juraska2019viggo,juraskawalker21}. 
We directly use the ViGGO corpus's  training, development and test sets. 
Each corpus instance uses one of Viggo's 9 dialogue acts 
such as \emph{verify attribute}, \emph{request explanation}, or \emph{recommend}. 
Most DAs are compatible with many combinations of content slots, using 14 video game attributes, yielding hundreds of response types \cite{juraskawalker21}. Figure~\ref{fig:viggo-dataset-examples} shows four DAs with various slot combinations.
 
\begin{figure}
    \small
   	\centering
    \begin{tabular}{p{4.5in}}
    	\hline
    	\cellcolor[gray]{0.9}
    	\emph{confirm}(\textsc{name} [\textbf{Hellblade: Senua's Sacrifice}], \textsc{release\_year} [\textbf{2017}], \textsc{developer} [\textbf{Ninja Theory}]) \\
    	Oh, do you mean the \textbf{2017} game from \textbf{Ninja Theory}, \textbf{Hellblade: Senua's Sacrifice}? \\
        \hline
        \cellcolor[gray]{0.9} \emph{suggest}(\textsc{name} [\textbf{Half-Life 2}], \textsc{genres} [\textbf{shooter}],
        {\textsc{play\-er\_per\-spec\-tive}} [\textbf{first person}] \\ 
        Do you also enjoy playing \textbf{first-person shooters}, such as {\bf Half-Life 2}?\\ \hline
    	\cellcolor[gray]{0.9}
    	\emph{give\_opinion}(\textsc{name} [\textbf{SpellForce 3}], \textsc{rating} [\textbf{poor}], \textsc{genres} [\textbf{real-time strategy, role-playing}], \textsc{play\-er\_per\-spec\-tive} [\textbf{bird view}]) \\
    	I think that \textbf{SpellForce 3} is \textbf{one of the worst games} I've ever played. Trying to combine the \textbf{real-time strategy} and \textbf{role-playing} genres just doesn't work, and the \textbf{bird's eye view} makes it near impossible to play. \\
        \hline
    	\cellcolor[gray]{0.9}
    	\emph{verify\_attribute}(\textsc{name} [\textbf{Little Big Adventure}], \textsc{rating} [\textbf{average}], \textsc{has\_multiplayer} [\textbf{no}], \textsc{platforms} [\textbf{PlayStation}]) \\
    	I recall that you were \textbf{not that fond} of \textbf{Little Big Adventure}. Does \textbf{single-player} gaming on the \textbf{PlayStation} quickly get boring for you? \\
        \hline
    \end{tabular}
	\caption{Viggo structured MRs (gray rows) and the corresponding reference utterances (with slot mentions in bold). Dialogue Act types are indicated in italics at the beginning of the MRs.}
    \label{fig:viggo-dataset-examples}
\end{figure}

\section{Experimental Setup}
\label{methods-sec}

We utilize the models GPT-Neo and Jurassic-1 jumbo
\cite{black2021gpt,lieberetal21}. GPT-Neo is a transformer-based
language model  that has 1.7 billion parameters. It was created as a open-sourced alternative to GPT-3. Similarly to previous GPT-2 and GPT-3 models, 
GPT-Neo predicts the next word given the previous words in the text. The team from EleutherAI 
generated an open source training set, The Pile \cite{gao2020pile}, comparable to that used for GPT models. 
The Pile is 825GB with data from 22 diverse sources, such as academic sources(Arxiv, PubMed), Github, and Wikipedia.  GPT-Neo  
has a vocabulary size of $\sim$50K tokens.
The EleutherAI team provides three  models (125M, 1.3B, 2.7B), which were trained as  masked auto-regressive models 
using cross-entropy loss. When compared  to the closest GPT-3 model (GPT-3 Ada), GPT-Neo 2.7B had better performance on all linguistic and scientific reasoning benchmarks (Hellaswag, Piqa, Winogrande, MathQA, PubMedQA). 
We use GPT-Neo 1.3B, which has promising performance for its size.\footnote{Experiments with GPT-2-small showed that models tuned with the $\sim$32K train did not
generalize  to unseen relations within the tuning domain, such as  from
the {\sc director} to the {\sc screen writer} relation, nor did these models generalize
across domains.}

\begin{wrapfigure}{r}{2.4in}
\centering
   	\vspace{-.2in}
    \begin{tabular}{p{2.3in}}
    	\hline
[{\sc prompt}]: confirm = yes | name = Tony Hawk's Pro Skater 3 | release\_year = 2001 | genres = sport \\ \hline
[{\sc sentence}]: Gotcha! So you're referring to the Tony Hawk's Pro Skater 3 sports game, which was released in 2001? \\ \hline
\end{tabular}
\vspace{-.05in}
\caption{Input format similar to QA}
\label{qa-format-fig}
\vspace{-.31in}
\end{wrapfigure}
Jurassic-1 is also an auto-regressive 
transformer-based language model, that
achieves state of the art performance on a set of common sense and QA
zero-shot and few-shot tasks \cite{lieberetal21,zellers2019hellaswag,sakaguchi2020winogrande}.
 AI21 Labs has released  two versions, J1-large with 7.5B parameters
and J1-jumbo with 178B parameters. Jurassic-1 is
pre-trained with 300B tokens taken from publicly available 
resources, and  has a larger vocabulary than other similar models
with 250K tokens. 
Jurassic-1 has a larger vocabulary by including n-gram phrases as tokens
along with the standard unigram and subword tokens. 
Jurassic-1's  architecture attempts to 
optimize the Jurassic's depth-width tradeoff \cite{levine2020depth,lieberetal21}. The paper claims that 
Jurassic-1 can predict text from a broader set of domains  than GPT-3,  and is superior to GPT-3 in few-shot settings,
due to its ability to fit more examples into a prompt.
We use temperature $=$ 0.7 
to promote interesting and varied output: the effect of temperature is illustrated in Figure~\ref{ai-studio-fig} by the multiple outputs. 

\begin{wrapfigure}{r}{2.6in}
\small
\centering
   	\vspace{-.2in}
    \begin{tabular}{p{2.55in}}
    	\hline
Starship = song = We Built This City | We Built This City = genre = pop rock\\
Starship plays pop rock like the song We Built This City.  Do you like that genre?\\ \\ 
Scream = cast member = Liev Schreiber \\
Liev Schreiber was really good in Scream, don't you agree?. \\ \\

name=Babbo | eatType = bistro | food = French | customerRating = outstanding \\ \hline
{\it- Babbo's bistro and restaurant serves French cuisine. The food is outstanding, according to customer reviews.}\\
{\it - Babbo is an outstanding French bistro in NY.  Do you like French food?}\\ 
{\it - Babbo is a bistro.  French food is the best.  Babbo has outstanding food.}\\
{\it - Babbo restaurant is rated outstanding. I heard they serve great food.} \\
\hline
\end{tabular}
\vspace{-.1in}
\caption{S2S Input format. Jurassic example, generating
multiple outputs (in italics) with no in-domain conditioning. }
\label{ai-studio-fig}
\vspace{-.2in}
\end{wrapfigure}
We have two large  datsets (Section~\ref{data-sec}), but we focus on GPT-Neo and Jurassic-1
few-shot (2, 3, 10) experiments, for two prompt formats, since format matters for Jurassic-1 \cite{lieberetal21}. 
In the QA format in Figure~\ref{qa-format-fig}, the prompt instances consist of input MRs marked as the {\sc prompt}
and the response marked as {\sc sentence}. 
The  S2S format in the top of Figure~\ref{ai-studio-fig} simply separates the MR and text into two lines, with an empty line separating test instances.  In Figure~\ref{ai-studio-fig}, 
the 2-shot conditioning prompts are from the  {\it music} and {\it movies} domains, and the test item is from the {\it restaurant} domain. We  generate multiple outputs, shown in italics, to  illustrate the effect of temperature.
All of the outputs are natural and coherent. Only the last output, {\it Babbo restaurant is rated outstanding. I heard they serve great food}, fails  to realize  all the MR attributes, missing eatType=bistro and food=French.

\begin{wrapfigure}{r}{2.8in}
    \vspace{-0.2in}
    \small
   \begin{tabular}{@{} p{.3in}|p{.4in}|p{2.0in}@{}} \toprule
        \bf ID & \bf Topic & \bf Novel Relations MR \\ \hline \hline
        M1& \bf Movies& (Despicable Me, screen writer, Cinco Paul) \\\hline
        M2& \bf Music& (The Beach Boys, song, Cotton Fields),  (Cotton Fields, date, 1970) \\\hline
        M3& \bf TV &  (Desperate Housewives, narrative location, Fairview)
 \\\hline
        M4& \bf Sports & (Muhammad Ali, significant event, lighting the Olympic cauldron),  (lighting the Olympic cauldron, of, 1996 Summer Olympics)
\\\hline
        
    \end{tabular}
    \caption{Sample Novel Test Meaning Representations used to test 2-shot prompt-based conditioning}
    \label{figure:athena-novel-MRs}
    \vspace{-0.2in}
\end{wrapfigure}
The 2-shot experiments  are intended to create a challenging task for testing the models' ability to generalize. In addition
to 2-shot conditioning with the two examples in Figure~\ref{ai-studio-fig}, the test set
consists of  novel hand-crafted  
MRs that are currently  not in Athena, which in some cases also use rare relations. The goal is
to  test  how well the models do at  realizing responses directly
from the WikiData KG, without any domain-specific or relation-specific conditioning. 
Table~\ref{ai-studio-fig} illustrates a good case of generalization to the
restaurant domain. 
Table~\ref{table:kg_topics} indicates with a * those
entities and relations corresponding to the novel MRs in our test set, and  example  novel MRs for
each topic domain are in Figure~\ref{figure:athena-novel-MRs}.

For evaluation metrics, we use {\sc bleurt}  along with human evaluation for the following metrics:
(1): {\bf coherence}: makes sense and is natural; (2) {\bf semantic accuracy}: triples realized divided by total triples
for the KG RGs and attributes realized divided by total attributes for Viggo; (3)
{\bf good  hallucinations}: additional {\bf true} information,  not specified in the MR, is added to the utterance
from the LM's own knowledge; (4) {\bf bad hallucinations}: additional {\bf false} information
is added to the utterance
from the LM's own knowledge; (5)  {\bf dialogue act accuracy}: whether the
output utterance matches the dialogue act specified  
for Viggo, exemplified in the
outputs in Figure~\ref{fig:viggo-dataset-examples}; (6) whether a question is added to  the end of the response, that was not specified in the MR or by the dialogue act, as seen in
the 2nd example output in Figure~\ref{ai-studio-fig}. Remember that  no dialogue acts are specified by the MRs for the Athena KG-RGs, but that
some of the Viggo dialogue acts, such as {\it suggest} typically are realized as questions
or include a question. 
For the 2-shot experiments with the novel  MRs, there are  no reference
utterances and {\sc bleurt}  scores cannot be calculated, so we use the human evaluation metrics. 

It is important to note that {\sc bleurt}  scores by themselves are not intended to mean anything: they are only  useful for  comparing  models \cite{sellam2020bleurt}.
In addition, {\sc bleurt}, like other n-gram scoring metrics, doesn't
account for stylistic variation
which is often desirable \cite{oraby2019curate,harrison2019maximizing}. Also, previous work  shows that the correlation of {\sc bleurt}  to human ratings of naturalness
varies across conversational
domains \cite{yeh2021comprehensive}. However,  that  work was based on crowd-sourced open-domain dialogues
where both sides of the dialogue were produced by humans. Here it might be expected
that {\sc bleurt} would be a good predictor of semantic accuracy. 
Therefore we use {\sc bleurt}  as  first
indicator of a model's performance and use {\sc bleurt}  scores to decide  whether
to perform human evaluation on a model's output. 
Then we 
examine whether the {\sc bleurt} scores are highly correlated
with  the human metrics for coherence and semantic accuracy.

\vspace{-.2in}
\section{Experimental Results}
\label{results-sec}


We report results for all the KG-RG topics and for Viggo, with both GPT-Neo and Jurassic-1.  The models were  also conditioned and tested for both the the QA format in Figure~\ref{qa-format-fig} and the S2S format in Figure~\ref{ai-studio-fig}. For the KG-RG topics, we also experiment with all possible cross-domain combinations of conditioning and test.

\begin{table}[h!tb]
\includegraphics[width=\textwidth]{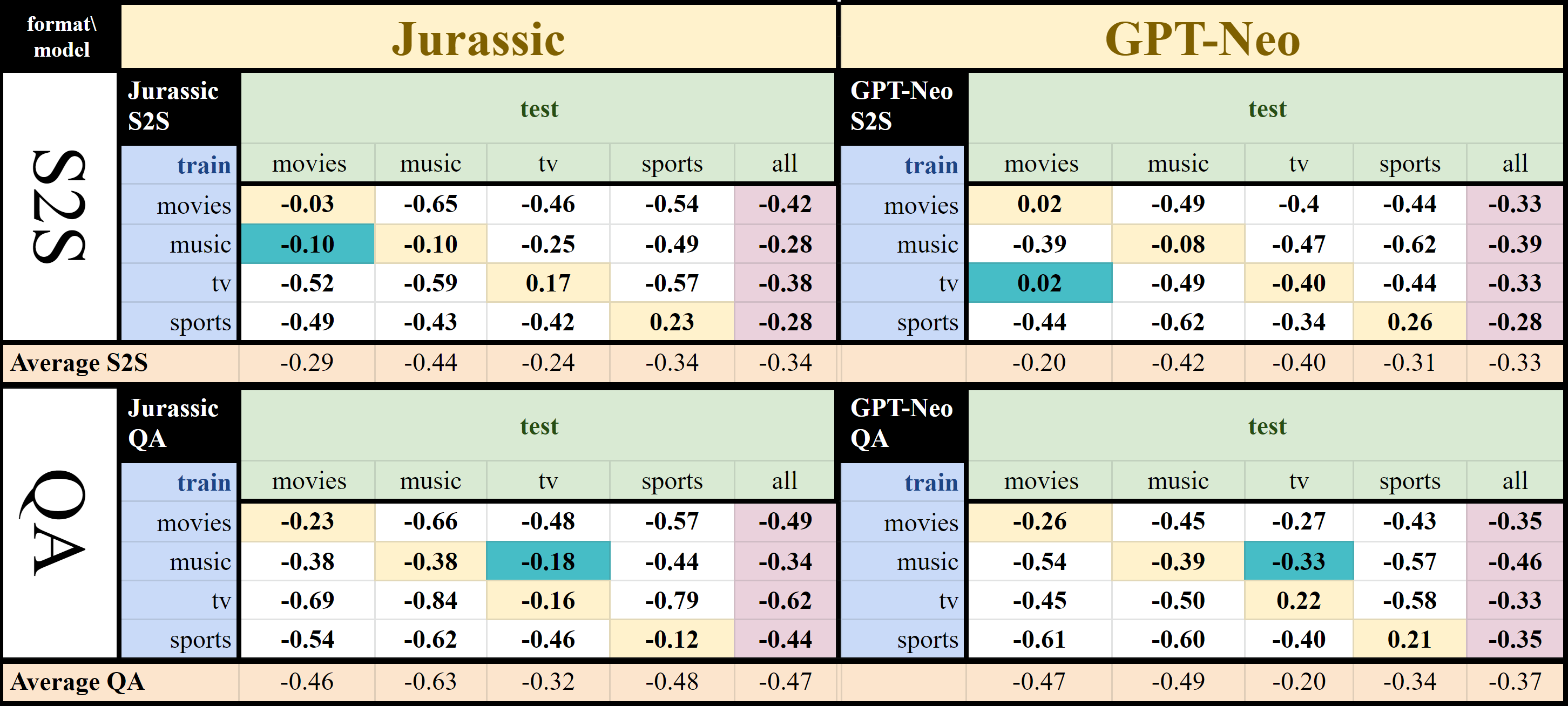}
\vspace{-.1in}
\caption{{\sc bleurt}  scores for testing within and across domain for Athena-Jurassic and Athena GPT-Neo. Prompt inputs in either S2S or QA format, conditioning on 10 instances of each topic.} \label{cecilia-pretty-table-10prompt}
\vspace{-.1in}
\end{table}

\vspace{0.1in}
\noindent{\bf Few-Shot Knowledge-Graph Response Generation.}
For each topic (movies, music, sports, TV), we randomly select ten instances for conditioning and 50 for testing (200 total). We tune Jurassic-1 and GPT-Neo with each conditioning set and then test each model on all four topics (test on 200) to examine both within and cross-domain few-shot performance. Table~\ref{cecilia-pretty-table-10prompt} provides the {\sc bleurt}  results for both Athena-GPT-Neo and Athena-Jurassic and for both S2S and QA formats. Rows indicate the conditioning domain, while
columns indicate test domains. The diagonal of each subtable
reports within-domain performance.  The average {\sc bleurt}  scores over all topics for each conditioning set are in the last column of each subtable, and averages for each input format (S2S or QA) are also included.

As expected, the within-domain results (highlighted in yellow) show that the models perform best when prompts are
from their own domain. The best results for in-domain conditioning are for sports, with an average {\sc bleurt} score of 0.23 for the S2S format for Jurassic, and 0.26 for the S2S format for GPT-Neo, as well as a 0.21 for the QA format for GPT-Neo. 
The within-domain performance for  the TV domain is also good, with a score of 0.22 for the QA format for GPT-Neo, and a score of 0.17 for Jurassic for the S2S format. Interestingly, sometimes a specific topic's prompts  perform as well or better for another topic than its own (highlighted in turquoise), e.g., GPT-NEO S2S conditioned  with TV prompts performs better on movies than TV, and Jurassic QA, when conditioned with music prompts, performs better for TV. This could arise because two domains are similar (TV and movies)  or because one domain is easier, e.g., the averages across the columns of each section suggest that TV is easier. 
 
The averages also clearly indicate that, for Jurassic, the S2S format works better, with large differences across all topic columns and topic diagonals, and an overall S2S of -0.34 compared to QA of -0.47 (p $<$ .01). For GPT-Neo, the overall differences between S2S (-0.33) and QA (-0.37) are not significant, and the story is more complex because  GPT-Neo QA works well for both TV (0.22) and sports (0.21). The differences between S2S and QA are not significant for TV or movies, but GPT-Neo S2S is significantly better than GPT-Neo QA for music and sports. 

A comparison of {\sc bleurt}  scores for  S2S for Jurassic vs. GPT-Neo for each  topic, shows that GPT-Neo is significantly better for Movies  (p $=$ .007), Jurassic is significantly better for music (p $=$.005), GPT-Neo shows a trend to be better for TV (p $=$ .07) and there are no differences for Sports (p $=$ .87). However, a paired t-test comparing {\sc bleurt} scores across all topics for both GPT-Neo and Jurassic shows that the overall differences are not significant. 

Since the overall differences for GPT-Neo  S2S are not significantly different than GPT-Neo QA, we focus the human evaluation on comparing  Athena-Jurassic to Athena-GPT-NEO for the S2S format. 
This will allow us to directly compare the human metrics for the two models
while the prompt format is fixed. We restrict the annotation to the within-domain testing. We sampled 30 of the 50 test examples
for each topic (240 examples). Three experts familiar with Athena labeled each output for coherence, semantic accuracy, good and bad extra information (hallucinations), and whether a question was added to the end of the response  (remember that no dialogue acts were specified in the Athena-KG  MRs). We also counted the number of words in each output to measure some aspects of the style of the outputs. 

\begin{wraptable}{r}{3.1in}
\includegraphics[width=3.1in]{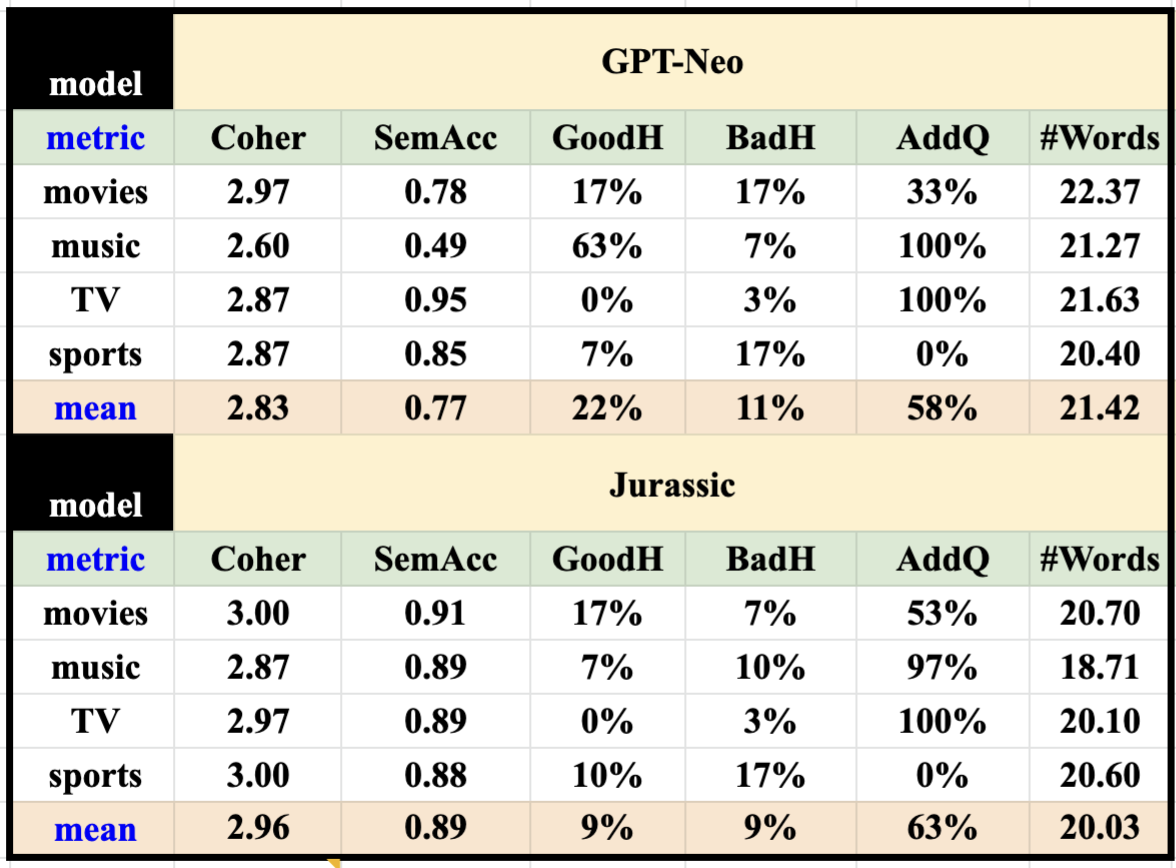}
\vspace{-.2in}
\caption{Human Metrics for  GPT-Neo  vs. Jurassic  per Topic. }
\label{athena-topics-human}
\vspace{-.2in}
\end{wraptable}
Table~\ref{athena-topics-human} presents the results for the human metrics,   showing that the average coherence ({\bf Coher}) for Athena-GPT-Neo is significantly lower than Athena-Jurassic (p = .002), as well as the semantic accuracy ({\bf SemAcc}) (p = .002). Athena-GPT-Neo hallucinates much more (total 33\%) than Athena-Jurassic (total 18\%). Although 22\% of Athena-GPT-Neo's hallucinations are good,  11\% of the utterances include at least one bad hallucination, in addition to failing to realize the semantic triples included in the MRs.

Both models frequently add questions, which can be considered as either a feature or a bug. They are usually quite natural, but the Athena KG MRs do not specify dialogue acts. Thus, there is no way to control whether the questions appear or not. The models use a similar number of words. Still, given the low semantic accuracy of Athena-GPT-Neo, many  words in the GPT-Neo outputs must be due to hallucinations.

The human evaluation suggests large differences between Athena-GPT-Neo and Athena-Jurassic in terms of the quality of the output, but there were no significant differences in overall {\sc bleurt} scores between GPT-Neo  S2S and Jurassic S2S. 
We therefore test whether {\sc bleurt} scores are meaningful proxies for 
the human metrics of coherence and semantic accuracy, finding that 
the correlation across all the topics, for each model indicate that {\sc bleurt} scores do  correlate with semantic accuracy,
with a correlation of 0.65 for Athena-GPT-Neo, and 0.46 for Athena-GPT-Jurassic. Coherence is  not correlated with {\sc bleurt}, with correlations of .02 for Athena-GPT-Neo and -0.05 for Athena-Jurassic.

\begin{figure}[t!hb]
    \centering
    \small
   \begin{tabular}{@{} p{.2in}|p{.4in}|p{1.7in}|p{1.9in}@{}} \toprule
        {\bf ID} & {\bf Topic} &  {\bf Input MR} & {\bf  Output}  \\ \hline 
        \rowcolor{light-gray}
        \multicolumn{4}{c}{\bf Athena-GPT-Neo} \\ \hline
      M1  &\bf Movies & (Peter Capaldi, award, BAFTA Award for Best Short Film),  (BAFTA Award for Best Short Film, show, 47th British Academy Film Awards),  (BAFTA Award for Best Short Film, work, Franz Kafka's It's a Wonderful Life) &	I think it's really great when a talented actor wins an award. do you think Peter Capaldi deserved to win a BAFTA Award for Best Short Film in 1980, for Franz Kafka's It's a Wonderful Life? \\ \hline
M2 &\bf Music &  (Kellie Pickler, song, Red High Heels),  (Red High Heels, genre, country music) &	Kellie Pickler is a country singer, and she's also a rapper. Do you know her songs?\\ \hline
\rowcolor{light-gray}
\multicolumn{4}{c}{\bf Athena-Jurassic} \\ \hline
M3 &\bf TV & (Saturday Night Live, award, Primetime Emmy Award for Outstanding Variety Sketch Series),  (Saturday Night Live, date, 2019)	&Saturday Night Live won a Primetime Emmy Award for Outstanding Variety Sketch Series in 2019. How does the fact that it got this award affect your opinion of the show? \\ \hline
M4 &\bf Sports & (Len Ford, member of sports team, Los Angeles Dons),  (Len Ford, position played on team, end) &	Did you know that Len Ford has played as a part of famous teams, such as the Los Angeles Dons, and played positions such as end.   \\\hline
    \end{tabular}
    \caption{Jurassic-1 and GPT-NEO outputs for conditioning with 10 within-domain prompts.}
    \label{tab:same-test-output}
    \vspace{-0.05in}
\end{figure}

Figure~\ref{tab:same-test-output} shows  examples for the two S2S models for each domain when tuned on within-domain prompts, which illustrate the strengths and weaknesses between models. 
The Athena-GPT-Neo output for M1 was labeled a 3 for coherence. However, it leaves out the triple (BAFTA Award for Best Short Film, show, 47th British Academy Film Awards).
It also includes the bad hallucination that Peter Capaldi is an actor, when in fact he wrote and directed the film.
In addition, the 47th British Academy Film Awards honored the best films of 1993, so Peter Capaldi won this award in 1994, not in 1980. The semantic accuracy annotation indicates that  2/3 triples are correct, the output includes two bad hallucinations, and  the output includes a question.
Similarly, the GPT-Neo output for M2 shows that GPT-Neo knows that Kellie Pickler is a rapper, knowledge that was not included in the MR. This was hand-annotated as a good hallucination. However, this output fails to realize the triple (Kellie Pickler, song, Red High Heels), so  semantic accuracy  was  1/2 triples. 
 
\begin{wrapfigure}{r}{2.6in}
    \centering
    \vspace{-.2in}
    \begin{small}
    \begin{tabular}{@{}|p{2.5in}|  @{}}
    \hline
    (Starship = song = We Built This City | We Built This City = genre = pop rock) \\
    Starship plays pop rock like the song We Built This City.  Do you like that genre? \\\hline
    (Planet of the Apes = cast member = Felix Silla)\\
    I heard Felix Silla starred in a good movie, called Planet of the Apes.\\ \hline
    \end{tabular}
    \end{small}
    \vspace{-.1in}
    \caption{Two prompt instances used with Jurassic-1 for 2-Shot Novel MR Experiments}
    \label{fig:jurassic-prompting}
    \vspace{-.2in}
\end{wrapfigure}
The Athena-Jurassic output for M3 was labeled as a 3 for coherence,
and that it includes a question.
The output correctly realizes all the triples so  it was marked as semantically perfect (3/3 triples realized). The output for M4 is also labeled
as a 3 for coherence. It  also correctly realizes all the triples (2/2), which are realized by a  {\it Did you know} question.
This output would not be annotated as including an additional question since the material in the {\it Did you know} question is part of the specified content in the MR.

\vspace{0.1in}
\noindent{\bf 2-Shot prompting on Novel Entities and Relations.}

\begin{wraptable}{r}{3.2in}
\vspace{-.25in}
\includegraphics[width=3.1in]{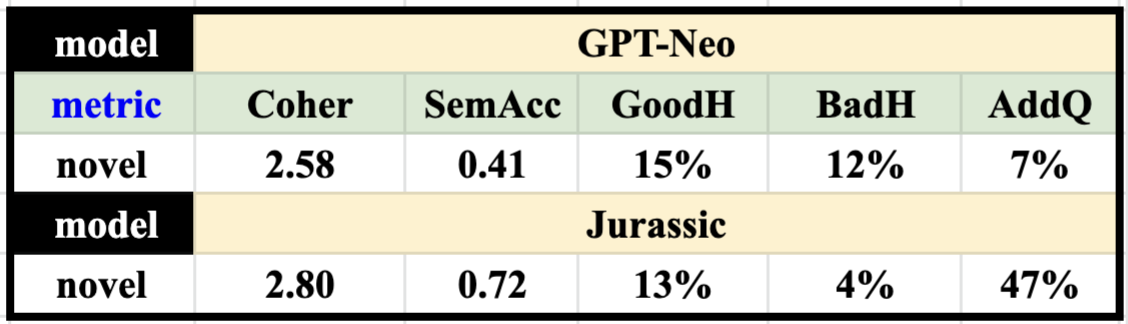}
\caption{Human evaluation for 2-shot Novel Athena-Jurassic vs. Athena-GPT-Neo, prompted with S2S format.} 
\label{athena-unseen-human}
\vspace{-.25in}
\end{wraptable}
\noindent We 
also performed  2-shot experiments  using the two prompt instances
for movies and music in Figure~\ref{fig:jurassic-prompting}. Because the realizations of each relation
or sets of relations requires a template to be written for Athena's current KG-RGs,
Athena has no templates for relations that are 
sparsely populated.   Thus, we test 
80 MRs composed of entities, relations, or combinations of
relations that are novel to Athena, as indicated
by a * in Table~\ref{table:kg_topics}. We only use 
the S2s prompt format since the results in Table~\ref{cecilia-pretty-table-10prompt}
show that the S2S format is clearly better for Jurassic and there are no
differences for GPT-Neo. 

\begin{figure}[t!h]
    \small
    \center
   \begin{tabular}{@{} p{.2in}|p{.4in}|p{1.8in}|p{1.8in}@{}} \toprule
      \bf ID & \bf Domain & \bf Input MR & \bf  Output  \\ \hline
       \rowcolor{light-gray} \multicolumn{4}{c}{\bf  Athena-GPT-Neo} \\
N1 & Movies & (The Black Cauldron, genre, film based on literature) &  
	The Black Cauldron is a film based on the novel of the same name by Stephen King. \\ \hline
N2 & Music  &  (Dua Lipa, award, Grammy Award for Best New Artist), (Grammy Award for Best New Artist, show, 61st Annual Grammy Awards)   
 &	Dua Lipa is a singer-songwriter from the Philippines. She is known for her hit songs ``I'm Not the One" and ``I'm Not the One (Remix)". 		\\ \hline
N3 & TV & (Psych, cast member, James Roday), (James Roday, character role, Shawn Spencer) & I heard James Roday starred in a good movie, called Psych.\\ \hline
N4& Sports &  (Tom Brady, award, Associated Press Athlete of the Year), (Associated Press Athlete of the Year, date, 2007) &	Tom Brady is the most valuable player in the NFL.  He is also the most valuable player in the NFL.    \\ \hline
\rowcolor{light-gray} \multicolumn{4}{c}{\bf Athena-Jurassic} \\
N5  & Movies  & (Ghost Rider, producer, David S. Goyer)  &
 Have you heard of a movie called Ghost Rider?  I heard it was produced by David S. Goyer.\\  \hline
N6   & Music & (Adele, award, Grammy Award for Song of the Year),  (Grammy Award for Song of the Year, show, 54th Annual Grammy Awards),  (Grammy Award for Song of the Year, work, Rolling in the Deep) & Adele won an award for Song of the Year at the 54th Annual Grammy Awards.  Do you like her song Rolling in the Deep? \\ \hline
N7 & TV & (Elementary, assessment, Bechdel test),  (Bechdel test, assessment outcome, passes) & Elementary passes the Bechdel test because at least two of the named women talk to each other about something other than a man. \\ \hline
N8    & Sports & (Usain Bolt, record held, Men's 100 metres world record progression),  (Usain Bolt, race time, 9.58 seconds),  (Usain Bolt, start time, 2009) & Usain Bolt holds the record for the men's 100 metres.  He broke the previous record, which was 9.69 seconds, at the 2009 World Championships in Berlin, Germany. \\ \hline
    \end{tabular}
    \caption{ Sample outputs for novel unseen relations and entities  for
    both Jurassic-1 and GPT-Neo}
    \label{tab:unseen-output}
    \vspace{-0.25in}
\end{figure}

Since there are no reference utterances for the novel MRs, {\sc bleurt} scores
cannot be calculated. Table~\ref{athena-unseen-human}  provides the results of the human evaluation, which indicate stark differences
between Athena-GPT-Neo and Athena-Jurassic, reflecting the reduction to 2-shot prompting, and the rareness
of the KG relations in the novel MRs.
The novel test MRs are also harder for both types of models, but 
the performance of Athena-GPT-Neo declines more sharply. Compare 
the mean coherence of 2.58 and semantic
accuracy of 0.41 for Athena-GPT-Neo in Table~\ref{athena-unseen-human} to the mean coherence of 2.83 and semantic accuracy of 0.77 for Athena-GPT-Neo
in Table~\ref{athena-topics-human}. Then compare  
the mean coherence of 2.80 and  semantic accuracy of 0.72 for 
Athena-Jurassic in  
Table~\ref{athena-unseen-human}, to the 
mean coherence of 2.96 and a semantic accuracy of 0.89 for 
Athena-Jurassic in Table~\ref{athena-topics-human}.
When testing with the novel MRs, Athena-GPT-Neo only adds
questions 7\% of the time, but Athena-Jurassic maintains a higher level and
adds questions 47\% of the time. When testing with the novel MRs. GPT-Neo  hallucinates bad information in 12\% of turns, while Jurassic only does so in 4\% of turns. 
This supports the claim by Lieber et al. that Jurassic should generalize better \cite{lieberetal21}.

Figure~\ref{tab:unseen-output}
provides  novel MRs  and outputs for all four topics for both Athena-GPT-Neo and Athena-Jurassic,
that illustrate the differences between the models. In N1, Athena-GPT-Neo produces a coherent
and semantically accurate utterance
about The Black Cauldron that includes a bad hallucination of Stephen King as the author,
when Lloyd Alexander is the author. In N2, 
Athena-GPT-Neo again produces a coherent utterance, but the content of that utterance doesn't include
{\bf any} of the triples in the MR, only matching  the name of the singer, Dua Lipa. In N3, the output is again
coherent, but it fails to realize the triple (James Roday, character role, Shawn Spencer). Example N4 illustrates
how GPT-Neo sometimes produces {\bf redundant} or {\bf logically inconsistent} outputs, where here it
says the same thing about Tom Brady twice, but sometimes it repeats itself many times, e.g.
{\it Friday Night Lights is a movie about a small town in Texas that is run by a family of criminals.  The town is run by a family of criminals. ...(4 times)}. In other cases, Athena-GPT-Neo contradicts itself.
There are  no examples from Athena-Jurassic that are redundant or logically inconsistent. In future work,   these categories could be added to the human metrics, even though they happen rarely. 

Figure~\ref{tab:unseen-output} also shows that Athena-Jurassic's  2-shot outputs are remarkably good.
In N5, the output is coherent, semantically correct and stylistically interesting. 
In N6, all three triples are realized correctly, and the last triple is embedded into a question, which
seems very natural. In N7, Athena-Jurassic  realizes all the content in the MR,
but also produces a good hallucination.
defining what the Bechdel tests actually is. 
  In N8, Athena-Jurassic seems to know a lot about Usain Bolt:  it
does not actually realize the triple (Usain Bolt, race time, 9.58 seconds), but provides the
race time for the previous record, and  produces a good hallucination of
the event that this happened at, namely the 2009 World Championships.

\vspace{0.1in}
\noindent{\bf  Few-Shot Response Generation for Viggo Video Games.} 
 We also experiment with few-shot prompt conditioning with the Viggo
corpus, with a  focus on the realization of dialogue acts. Athena KG MRs do not specify the
dialogue act, and thus its use of questions cannot be controlled. The dialogue acts in Viggo are shown in Figure~\ref{viggo-conv}
and  Figure~\ref{fig:viggo-dataset-examples}. The Viggo experiments compare
prompt conditioning  with  GPT-Neo and Jurassic, for both S2S and QA formats,
and compares 3-shot conditioning to 10-shot conditioning per dialogue act. All experiments use
a randomly selected set of 100 Viggo test items.


\begin{wraptable}{r}{2.9in}
\vspace{-.25in}
\includegraphics[width=2.8in]{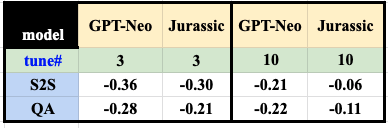}
\caption{{\sc bleurt}  for Viggo comparing GPT-Neo and Jurassic, S2S format vs. QA format, and
3 prompting instances vs. 10.} \label{viggo-BLEURT-table}
\vspace{-.3in}
\end{wraptable}
\noindent
Table\ref{viggo-BLEURT-table} provides the {\sc bleurt} scores for these prompting variations.
The QA row in Table \ref{viggo-BLEURT-table} for 3-shot conditioning suggests that tthe QA format performs  better for 3-shot than the S2S format. However for  10-shot conditioning, S2S is better for both GPT-Neo and Jurassic. The {\sc bleurt} scores for Athena-Jurassic
for 10-shot conditioning are clearly much better than any of the other results.

\begin{wraptable}{r}{3.0in}
\includegraphics[width=3.0in]{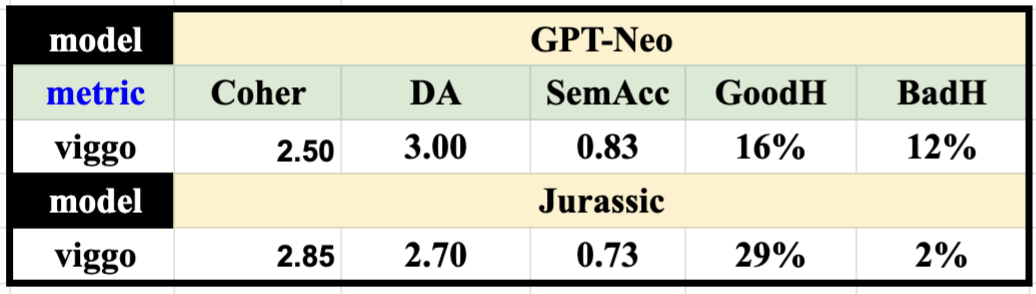}
\caption{Human metrics for Viggo comparing GPT-Neo and Jurassic} 
\label{viggo-human-table}
\vspace{-.25in}
\end{wraptable}
Table~\ref{viggo-human-table} provides the results of the human
evaluation showing that Athena-GPT-Neo does better at matching the dialogue acts (DA) and achieves higher semantic accuracy (0.83) than Athena-Jurassic. However, overall the results are good for controlling dialogue acts.
Table~\ref{viggo-human-table} also shows however that the coherence of Athena-GPT-Neo is poor compared to Athena-Jurassic. Athena-Jurassic demonstrates the knowledge that Jurassic-1 has about video games  by producing far
more good hallucinations (29\%), 
while only producing 2\% of bad hallucinations, as compared to the 12\% of bad hallucinations by Athena-GPT-Neo.

\begin{figure}[t!h]
    \small
    \center
   \begin{tabular}{@{} p{.2in}|p{2.2in}|p{2.2in}@{}} \toprule
      {\bf ID} & {\bf Input MR} & {\bf  Output}  \\ \hline
      \rowcolor{light-gray} \multicolumn{3}{c}{\bf  Athena-GPT-Neo} \\
V1 & confirm = yes, name = Tony Hawk's Pro Skater 3, release\_year = 2001, genres = sport
 &You mean the Tony Hawk's Pro Skater 3 that has got a sport genre?   
	 \\ \hline
V2 &  give\_opinion = yes, name = Might \& Magic: Heroes VI, rating = average, player\_perspective = bird view, platforms = PC &
	Might \& Magic: Heroes VI is a solid game. I like that it has a multiplayer and, since it's only rated E (for Everyone), I can play it with friends and younger siblings.
 		\\ \hline
 \rowcolor{light-gray} \multicolumn{3}{c}{\bf Athena-Jurassic} \\

V3   &   confirm = yes, name = Tony Hawk's Pro Skater 3, release\_year = 2001, genres = sport &
	Gotcha! So you're referring to the Tony Hawk's Pro Skater 3 sports game, which was released in 2001? 
 \\  \hline

V4   &  give\_opinion = yes, name = Sid Meier's Civilization V, rating = good, available\_on\_steam = yes, has\_linux\_release = yes &
4X games can be really fun to play, and Civilization V is a pretty good example. It also happens to be available on Linux, and that's a huge plus for me.   \\ \hline
 \end{tabular}
    \caption{Viggo outputs with 10-shot conditioning per dialogue act for QA prompt format}
    \label{test-viggo-examples}
    \end{figure}

Sample outputs for both Viggo models are shown in Figure~\ref{test-viggo-examples}. Both models produce good outputs
much of the time.  The low coherence of Athena-GPT-Neo primarily arises from redundancy of repeating the same information twice. Compare example V1 to example V3. Both models get the dialogue act of {\it confirm} correct.
However, the Athena-Jurassic output in V3 is more varied and interesting and
includes all the attributes specified in the MR. The Athena-GPT output in V1 misses out the release year attribute,
and uses an odd phrasing {\it that has got a sports genre} that results in a lower coherence score. Example V2 illustrates  Athena-GPT-Neo's   knowledge about video games, with a good hallucination that it is
rated E for Everyone, and that means that it can be played with younger siblings. Example V4 illustrates
Athena-Jurassic failing to realize some of the MR attributes, such as the availability on Steam and the full
name of the game. However, the language is again very natural, e.g. {\it that's a huge plus for me}.


\section{Conclusion}
\label{conc-sec}

We used prompt-based learning to create new neural models for semantically-controlled meaning-to-text (M2T) natural language generators (NLGs) to improve the quality and the coverage of the M2T 
response generators in Athena, an open-domain dialogue system that has been a finalist in the Alexa Prize for the last two years~\cite{harrison2020athena,patilathena2021}. A major challenge for such  systems is the need to  produce truthful, high-quality responses on any topic. 
We created Athena-GPT-Neo and Athena-Jurassic  using GPT-Neo \cite{black2021gpt},
and Jurassic-1 \cite{lieberetal21}, by experimenting  with few-shot (2, 3, 10) prompt-based learning 
for Athena's knowledge-graph domains of movies, music, TV, sports and 
with the Viggo corpus's dialogue act-based MRs for  video games.  We also experimented with multiple prompt formats and 
with testing both within and across-domain. The ability to create NLGs that generate high-quality responses directly from MRs via few-shot prompt conditioning  will greatly facilitate the use of M2T NLGs in dialogue systems.
To our knowledge, these are the first results demonstrating  that few-shot prompt-based learning can create M2T NLGs that generalize well to new semantic domains.

Athena-Jurassic produces high-quality, semantically-controlled, conversational responses directly from MRs and KG triples. These results confirm the choice that the Jurassic-1 creators made
to use a larger vocabulary with phrasal tokens, and 
less depth and more width, in order to  create a  model that  generalizes better  \cite{lieberetal21,levine2020depth}. Our results show that both Athena-GPT-Neo and Athena-Jurassic generally produce coherent 
output with  10-shot within-domain conditioning, 
but that Athena-Jurassic is significantly better for both coherence and semantic
accuracy.  While
we have not tested whether real-time response generation is possible, we believe 
 the responses are generally of high enough quality to be used in settings with real human users,
 such as the Alexa Prize \cite{gabriel2020further,Venkatesh2017,patilathena2021,harrison2020athena}.
 We plan to do additional experiments with Viggo in order to improve its performance to the level
 required \cite{juraskawalker21}.
 
We also showed that Athena-Jurassic  performs well 
with 2-shot conditioning, using completely  novel  sets of KG triples with unseen relations
and entities. These novel MRs are not currently 
included in Athena, because the relations are rare, and creating templates
for novel relations or sets of relations is typically not worth the human effort \cite{rambow2001evaluating,rambow2001natural}.
For example the MR in M4 in Figure~\ref{figure:athena-novel-MRs} describes
the event of Muhammed Ali lighting the Olympic torch in 1996, a rarely populated
event for the athlete entity type. Athena-Jurassic achieves a semantic accuracy
of 2.72 out of 3 for MRs like this in our challenging 2-shot  setting.

In experiments with the KG  response generators in Athena, we found that in almost
half the responses, Athena-Jurassic adds questions to the end of the response, which are
typically quite natural. However the use of questions cannot
be controlled because the  KG-RG meaning representations do not specify dialogue acts. Thus 
we also experimented with  few-shot conditioning for controlling dialogue acts using the MRs in the Viggo video games corpus.  We 
showed that both Athena-GPT-Neo and Athena-Jurassic can learn to control dialogue
acts with 10-shot conditioning per dialogue act. However again, Athena-Jurassic performs significantly
better on the human metrics of coherence and semantic accuracy. Interestingly, often Athena-GPT-Neo
successfully produces the {\it form} or {\it syntax} of the dialogue act, e.g.   a verify-attribute
dialogue act,  while getting very few of the MR attributes correct.
For example, Athena-GPT-Neo produces
{\it 	You said you liked Assassin's Creed Chronicles: India. Do you think it would have been better to make it a single-player only game?} when the reference utterance is {\it 	So I know you said you hated Assassin's Creed Chronicles: India. Do you think all of Climax Studios side view games are as bad?}.
Here, Athena-GPT-Neo
only gets the name attribute correct, and misses the attributes that it is single-player, the user-rating is poor, and
the developer is Climax Studios. 

We also presented automatic evaluation results using {\sc bleurt} for cross-domain
testing. Some of the {\sc bleurt} results are very good, and suggest that
cross-domain 10-shot conditioning can also produce high quality utterances.
Our results also show that {\sc bleurt} scores have  good correlation with the human
metric of semantic accuracy, but not coherence. Future work should evaluate 
these cross-domain results with human metrics. It would also be valuable
to experiment with the large number of recently
proposed automatic evaluation metrics to test whether there are better metrics
than {\sc bleurt}
for doing automatic evaluation in this task setting \cite{howcroft2020twenty,yeh2021comprehensive}.
Many recently proposed automatic metrics rely on evaluating outputs within
a dialogue context, which typically is not available in M2T NLG experiments.
However there are also novel reference free metrics that could be tested in this setting.

There are many other possibilities with both the WikiData knowledge graph RGs and with corpora such as
Viggo for prompt-based learning and testing regimes that we have not yet experimented with or fully evaluated.  We also plan to carry out future experiments  on a number of other challenging
problems for NLG \cite{reedetal20,oraby2019curate,oraby2018neural,harrison2019maximizing}.


\end{document}